# Single-sample writers – "Document Filter" and their impacts on writer identification

**Fabio Pinhelli**
Department of Informatics
State University of Maringá
Maringá - Paraná - Brazil
fabio.pinhelli@gmail.com

Alceu S. Britto Jr.
Pontifical Catholic
University of Paraná
Curitiba - Paraná - Brazil
alceu@ppgia.pucpr.br

Luiz S. Oliveira
Federal University of Paraná
Curitiba - Paraná - Brazil
luiz.oliveira@ufpr.br

Yandre M. G. Costa
Department of Informatics
State University of Maringá
Maringá - Paraná - Brazil
yandre@din.uem.br

Diego Bertolini
Federal Technological
University of Paraná
Campo Mourão - Paraná - Brazil
diegobertolini@utfpr.edu.br

May 14, 2020

## Abstract

The writing can be used as an important biometric modality which allows to unequivocally identify an individual. It happens because the writing of two different persons present differences that can be explored both in terms of graphometric properties or even by addressing the manuscript as a digital image, taking into account the use of image processing techniques that can properly capture different visual attributes of the image (e.g. texture). In this work, perform a detailed study in which we dissect whether or not the use of a database with only a single sample taken from some writers may skew the results obtained in the experimental protocol. In this sense, we propose here what we call "document filter". The "document filter" protocol is supposed to be used as a preprocessing technique, such a way that all the data taken from fragments of the same document must be placed either into the training or into the test set. The rationale behind it, is that the classifier must capture the features from the writer itself, and not features regarding other particularities which could affect the writing in a specific document (i.e. emotional state of the writer, pen used, paper type, and etc.). By analyzing the literature, one can find several works dealing the writer identification problem. However, the performance of the writer identification systems must be evaluated also taking into account the occurrence of writer volunteers who contributed with a single sample during the creation of the manuscript databases. To address the open issue investigated here, a comprehensive set of experiments was performed on the IAM, BFL and CVL databases. They have shown that, in the most extreme case, the recognition rate obtained using the "document filter" protocol drops from 81.80% to 50.37%.

Signature verification and writer identification, Document analysis systems, Biometric systems and applications.

## 1 Introduction

Writer identification is an ordinary task, regularly necessary in some specific application domains, such as forensic science for example. In this kind of application domain, the need for an accurate identification of the person who wrote a given manuscript is a recurrent problem. The automatic writer identification is a hot topic, intensively addressed by the pattern recognition research community, which has raised from scenarios like this.

Another important fact that has contributed to the increasing of this field of research is the growth of the amount of manuscript documents stored as digital images. Moreover, the development of biometric systems has been the subject of investigations of several researchers [18]. This kind of system has shown to be capable of supporting experts on decision-making in a wide range of areas.

Before going further into details about the specific goals and methods of this work, it is important to clarify some general concepts that are closely related to this subject of investigation. The first of these concepts that we would describe here concerns to the notion of *online* and *offline* handwritten identification techniques. The *online* techniques are those in which a lot of additional information, regarding the manuscript writing is available, such as the writing speed, the pen pressure, and the direction of the writing. This kind of technique depends on the electronic device used to collect the information (e.g. graphic tablet). The *offline* techniques, on the other hand, are those in which all the information available is the manuscript image, and only this. In many cases, these images are obtained by scanning a manuscript already written on a paper [3]. Despite of the fact that the *offline* techniques tend to be more affected by limitation issues, most of the investigations are developed in this scenario because most of the databases available to the community research are created on the *offline* mode, and this is the way on how the manuscripts are available in the vast majority of the real problems.

The second aspect regards the manuscript databases, which can be classified as text-dependent and text-independent.Text-dependent are those in which the volunteers are asked to copy the content of previously defined documents. By this way, all the volunteers are supposed to produce manuscripts with approximately the same size (i.e., same amount of words). On the opposite direction, text-independent are those in which the volunteers can freely describe a manuscript, without limits that would enforce the production of a certain quantity of text. In general, the use of text-independent produces a much more challenging scenario, because there is no warranty that each writer has produced an enough amount of text which ensures the creation of good classifiers models during the training step [3].

Another important categorization concerns the task itself. According to Jain et al. [9], there are three kinds of tasks commonly addressed on this field of study and they are closely related to the intended objectives:

- **Writer verification**: (Is the writer who he claims to be?). In this kind of task, the system performs a comparison between the pattern taken from the supposed author of the manuscript and the queried document. It corresponds to an one-to-one search (1:1);
- **Writer identification**: (Who is the writer?). Given a sample, the system performs a search among all the patterns present in the database trying to find the pattern which better corresponds to the queried document. It corresponds to an one-to-many search (1:$N$);
- **Writer monitoring**: (Is the writer a wanted person?). Generally used in airports and surveillance systems, aiming to detect if a given handwritten manuscript corresponds to the manuscript of a wanted person, enrolled in a list. It corresponds to an one-to-many search (1:$k$, in which $k$ is the size of the list of wanted persons).

Once we have described some important concepts regarding this field of investigation, we will use in this work the proper terminology, already introduced. This paper describes a set of experiments performed to accomplish writer identification on the *offline* scenario by using two documents from each writer on the IAM [13], BFL [6] and CVL [11]. The main contribution of this work is the introduction of the "Document Filter" (DF) protocol, which enforces that all the data representation taken from fragments of the same document must be placed exclusively either in the training or in the testing set. A comprehensive set of experiments was carried out, and they confirm that, in the most extreme cases, the identification rate obtained using the DF protocol can drop from 83.6% to 69.7% using IAM, 81.8% to 50.3% with BFL, 76.5% to 62.6% in the CVL database.

This paper is organized as follows: in Section 2 we describe the motivations that encouraged us to develop this work; Section 3 presents details about the experimental setup; Section 4 shows the obtained results; a critical review is described in Section 5; and finally we present the concluding remarks.

## 2 Motivation

Several works already presented on the literature describing methods intended to address the writer identification (or writer verification) task are developed using databases created on the text-dependent mode. Furthermore many of these databases have just a single manuscript sample for great part of the volunteers who contributed to their creation. In one hand, the text-dependent mode tends to make easier the achievement of good results, once it ensures the availability of a reasonable amount of manuscript text for each writer to the development of the experimental protocol. On the other hand, researchers are enforced to use data taken from different fragments of the same manuscript both in the training and testing sets in the case of writers from whom we have just one document sample in the database. Taking these into consideration, the performance of several results already published in this field of research may be positively or negatively affected by conditions artificially introduced by limitations of the database in the accomplishment of these investigations.

There are several factors that can help to understand why using a single document from a writer can somehow introduce a bias to the results. Regarding to this matter, we can cite the emotional state of the writer; the fluency of the pen/pencil used for writing; and the paper writability (smooth/rough), among others [12].

By analyzing the literature, we can easily find several experiments attacking the problem investigated here with very high performance rates. However, as we know, in real world problems the scenery found is frequently different from that enforced by many databases used in the experimental setup. Usually, there is no plenty of manuscript samples per writer to build a more robust classifier model.

With that in mind, in this work we try to open up this "black-box" developing a series of experiments to evaluate a quite subtle issue surrounding the writer identification task aforementioned. For this purpose, we evaluate the impacts of the introduction of a mandatory restriction which prohibit the use of data taken from different fragments of the same manuscript both on the training and testing sets. We call this restriction "Document Filter" (DF), and it was inspired by a similar protocol introduced by Pampalk et al. [16] in the music genre classification task. In that work, the authors intended to avoid the creation of classifiers able to classify artist ("Artist Filter"), instead of music genre. It was reported that in the most extreme case, the recognition rate reduced from 71% to 27% when the filter was applied.

In the context of the DF assessment, here we decompose the IAM, BFL and CVL manuscript databases in such a way that all the data obtained from the same manuscript must be placed either in the training set or in the test set. Thus, we evaluate the differences on performances considering or not this filter. Hence, we have two different subsets: i) with samples from the same document in both training and testing sets; and ii) with samples from one document in the training set and from another document in the testing set.

### 2.1 Challenges

Writer identification and writer verification may be considered quite challenging tasks, because each writer corresponds to one class in these problems, thus characterizing a problem with a large number of classes. In addition, sometimes the writing style of different people present a reasonable similarity, what makes the inter-class likeness increase proportionally to the number of writers [17].

Another adversity is given by the fact that as well as two distinct people do not have identical manuscripts, one person does not reproduce its own manuscript twice identically [12]. This subtlety implies a difficulty to correctly identify at a glance either or not two manuscripts was made by the same person.

Another interesting question rely on the amount of text needed to build a classification model which properly generalizes the classes. Is it possible to perform the writer identification by creating the model using only one document sample per writer? And only a single text line per writer? And just a single word per writer? Questions like these still remain open for further investigations.

## 3 Experimental setup

In this section we describe the databases used to perform the experiments in subsection 3.1. Following, we describe in details how we organized the data aiming to get the desired evaluation in subsection 3.2. Lastly, we also present some information about the protocol used to perform the identification of writers in subsection 3.3.

### 3.1 Databases

The IAM database is composed of documents written in English and it presents a great variation regarding the amount of content per manuscript, since it has been created on the text-independent mode. The first version of the IAM, created in October 2002, was described by Marti and Bunke [13] and it has 115,320 samples of handwritten words, distributed in 13,353 lines of text. The handwritten content was produced by approximately 400 writers, and it has a lexicon with 10,841 different words. The number of handwritten samples collected per writer vary widely (i.e. from 1 to 59 samples per writer), and the vast majority (356 writers) contributed with only a single sample. Table 1 shows the distribution of documents per writer on the IAM database. Currently, the IAM 3.0 database contains samples taken from 657 writers and it is available online. This last one is the IAM version chosen to be used in this work.

The BFL database, proposed by Freitas et al. [6] is composed of manuscripts taken from 315 writers, and contains three document samples per writer (text-dependent). An interesting peculiarity of this database consists on the fact that the data collection was made in three separate sections, in different days. The text, wrote in Portuguese, was carefully choosen, such a way that it contains all letters, numbers, and special characters from the Portuguese Language. The volunteers who contributed to the creation of the database used her own pen, and the text was wrote in a white paper with no pen-draw baseline. Lastly, the documents were scanned in gray levels with a resolution of 300 dpi.

Table 1: Distribution of documents per writer on the IAM database.

| # Docs | 1 | 2 | 3 | 4 | 5 | 6 | 7 | 8 | 9 | 10 | 59 |
|---|---|---|---|---|---|---|---|---|---|---|---|
| Writers with # docs or more | 657 | 301 | 159 | 127 | 93 | 39 | 37 | 33 | 31 | 13 | 1 |
| Writers with # docs exactly | 356 | 142 | 32 | 34 | 54 | 2 | 4 | 2 | 18 | 12 | 1 |

The CVL database [11] is a text-dependent database, with contributions of 310 writers. The texts choosen are excerpts from literary works, and they contain from 47 to 90 words. This database can be employed in multi-script tasks as well, because it also contains documents wrote in English and German. Summing up 1,604 handwriting samples. In this database, all writers contributed with at least four documents in English and one in German. The document was digitalized generating images described using the RGB color space, with 300 dpi of resolution.

These databases were chosen to be used in this study because they have suitable characteristics to address the issues we intend to investigate. Moreover, they have been widely used in other works described in the literature.

### 3.2 Data organization

Taking into account that for some writers there is only one manuscript available in the databases considered, we divided each document in nine blocks with size $m \times n$. By this way, we have obtained more than one sample from these writers, what is mandatory to perform training and testing considering these writers as classes.

Firstly, we have analyzed the usage of all writers of the IAM database, but using only a single manuscript sample from each one, even for those who have more than one sample in the original database. In order to do that, we took the first document from each writer and set six from the nine original blocks to compose the training set, and the other three remaining blocks were taken as members of the testing set. We call the subset used in this experiment $IAM_{all}$[1], once we are using all the writers present in the database (i.e., 657 writers), although we are not using all the documents from each of them. We have done it three times (three folds): using the first six blocks, the first three and last three blocks, and the last six blocks as training sets, respectively. The results we show in the next section correspond to the mean of these three executions. Since we are dealing with a single document per writer, both the training and testing sets are composed of samples obtained from the same document. We call this "without DF".

In the second experiment, we intended to evaluated the performance obtained using manuscripts from all the 657 writers of the database, but using the DF protocol only for those writers who have more than one manuscript sample. In the case of writers from whom we have more than one document available (i.e. 301 writers), we also set three testing folds using blocks from the second document. We call this "with DF". It is worth to mention that in the case of single-sample writers, we kept the blocks from the same document once it was the only option in order to use the all writers of the database. We have used the same classification model in both cases. We only vary the testing set with blocks from the same document (i.e. "without DF") or not (i.e. "with DF").

After that, we isolated the 356 single-sample writers and performed the same classification scheme to evaluate only the behaviour of the data, which it is not possible to make investigations considering the DF protocol. These experiments were performed trying to get a better view of the influence of this part of the database on the results performed with the database as a whole. We call this subset $IAM_{v1}$[2], once we are using strictly the single-sample writers.

To analyze the impact of the DF protocol, we set a scenario in which the filter can be applied to every writer. For this purpose, we selected the first two documents from each writer with two or more documents (i.e. 301 writers). In this way, the DF protocol already described can be suitably applied to these data. We call this subset $IAM_{v2}$[3], once we took two documents from each writer with two or more documents.

---

[1]$IAM_{all}$: Dataset composed of only one manuscript sample from all the 657 writers who contributed to the creation of the database. Should not be confused with the full database.

[2]$IAM_{v1}$: Corresponds to the part of IAM database in which the writers (i.e., 356 writers) contributed with only a single manuscript sample to its creation.

In summary, to put the different ways of organizing the data described in this section according to the DF concept previously described, we are considering "without DF" those executions where the blocks in training and testing sets belongs to the same document (i.e., the scenario where the writer has only one document in the database), and we call "with DF" those executions where we consider the testing set as belonging to another document.

We have used essentially the same protocol used on the $IAM_{v2}$ subset (i.e., the IAM database with two or more documents) for the BFL [6] and CVL [11] databases.

### 3.3 Experimental protocol

In this work, the Speed Up Robust Features (SURF) was employed as the feature descriptor to be applied on the manuscript images. The idea behind SURF is to extract key points from different regions from a given image. This descriptor has been successfully used in different pattern recognition tasks. The key points are detected based on hessian matrix, and right after a square region is constructed around each key point identified. Following, the SURF generates a 257-dimensional feature vector, that is formed by concatenating the information from the sub-regions around each key point. More details about SURF can be found in [1].

Color and shape are among the most intuitive attributes explored in works which aim to capture the visual content of images. However, the textural content of images has also been successfully captured and used in several different pattern recognition tasks. The descriptors created to pick up the textural content of images, in general, capture information regarding their softness, roughness, and regularity among others [7].

Hanusiak et al. [8] proposed a method which allows to generate a visual representation of a given manuscript with its textural content emphasized. The "texture generation" process consists of the rearrangement of the connected components of the image, preserving the original writing inclination, but reducing spaces between lines, words and characters present in the text. As a result, it is generated an image that retains the characteristics regarding the writing style, but with a more dense textural content. The obtained image is quite suitable to be addressed using texture operators described in the image processing literature.

Figure 1(a) shows a block taken from the original image of a manuscript, in which we apply the SURF descriptor. Figure 1(b) shows a texture block generated from the orignal image, as proposed by Hanusiak et al. [8].

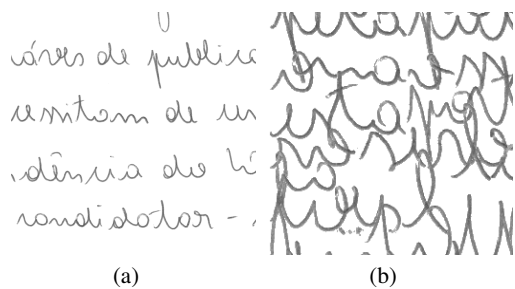

(a)                               (b)

Figure 1: Example of original block and texture block used in this work.

In this work, we have also used the texture generation protocol described in the preceding paragraph. In this sense, we chose the Local Binary Pattern (LBP) and the Local Phase Quantization (LPQ) texture operators. The LBP calculates a binary pattern for each pixel present in it. This pattern is found taking into account the grey level of the pixel and other pixels in its vicinity, according to some predefined criteria. Finally, the LBP feature vector corresponds to the histogram of the binary patters found for every pixel of the image. More details about LBP can be found in [14]. The LPQ texture descriptor, on the other hand, is base on the phase spectra of Short-Time Fourier Transform (STFT). It was originally proposed to be used on blurred image, but it also present a good performance on images not affected by that kind of noise. More details about this operator can be found in [15].

Support Vector Machine (SVM) was chosen as the classifier model to be used to perform classification here because it has already been successfully used in this specific task in several works described in the literature [3, 2, 18, 5]. The training and testing sets already described are classified using SVM with its hyper-parameters configuration chosen by GridSearch, using the widely-known LibSVM[4] library. Once the SVM predictions are obtained for each block individually, we need to perform a fusion of these predictions in order to get the final decision for the document as a whole. In this sense, we used the Sum fusion rule[10], since it obtained the best performance in the work [3].

---

[3]$IAM_{v2}$: a dataset created taking two documents from each one of the 301 writers who contributed with more than one manuscript sample.

As one can observe in the experimental results, presented in Section 4, in many cases, the reported results may appear lower than those already published in the literature on a similar scenario. However, in this paper we are not pursuing the improvement of the performance in terms of correct identification. Actually, we have introduced some additional restrictions to properly isolate the specific issue that we want to scrutinize here. Therefore, the reduction of performance is something expected in many cases. Figure 2 illustrates the proposed method. Figure 2(a) describes the usage on the "without DF" mode, and Figure 2(b) describes the usage on the "with DF" mode.

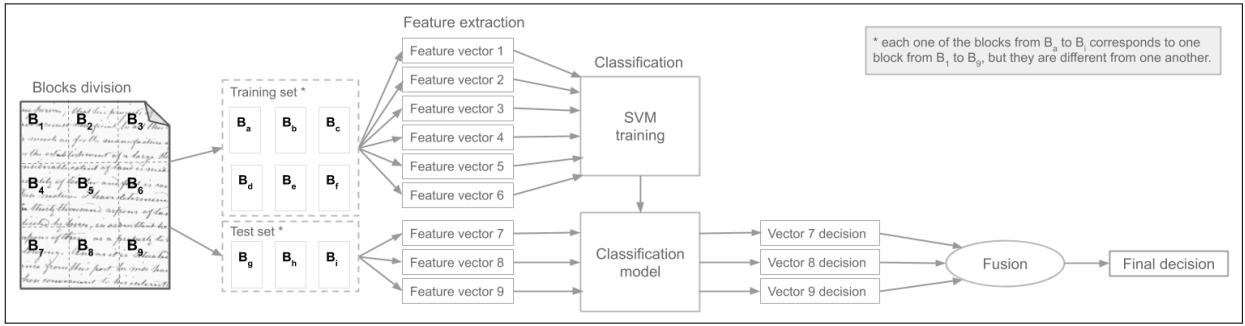

((a)) Evaluated scheme without the use of DF.

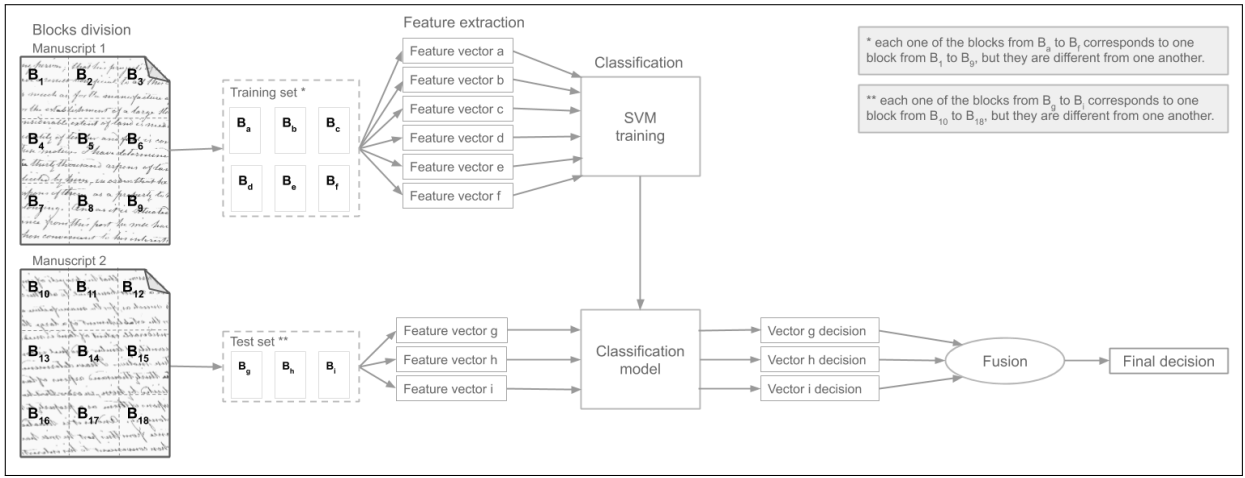

((b)) Proposed scheme using DF restriction.

Figure 2: General overview of the proposed/evaluated approach.

## 4 Experimental results

In this section we describe the results obtained in the experiments carried out on the IAM, BFL, and CVL databases.

Our main objective with this work is to evaluate whether the lack of use of the DF protocol can lead to misleading results. For this, we choose the three aforementioned databases, and three different feature descriptors (SURF, LBP, and LPQ). Tables 2, 3, and 4 describe the results achieved on these databases using these features.

The results described in Table 2 show us that when the SURF descriptor is used with the DF protocol, the performance rates decrease if compared to the rates obtained without the use of DF in all the evaluated databases. These results evidenciate that the use of data taken from the same manuscript simultaneously both on training and test sets tends to bias positively the results. In the most extreme case, the difference between the rates obtained with or without the use of DF reaches 13.95%, for the $IAM_{v2}$ database.

Table 2: Identification rates (%) using SURF descriptor

| Database | # Writers | Without DF ($\sigma$) | With DF ($\sigma$) | DIF |
|---|---|---|---|---|
| $IAM_{v2}$ | 301 | 83.67 (±4.62) | 69.71 (±3.94) | 13.95 |
| BFL | 315 | 51.53 (±4.54) | 43.81 (±2.26) | 7.7 |
| CVL | 310 | 76.56 (±7.03) | 62.62 (±8.33) | 13.93 |

Tables 3 and 4 describe the identification rates obtained using LBP and LPQ texture descriptors applied on texture blocks. As happened with the SURF descriptor, the use of the DF protocol has also reduced the performance rate on all databases. This leads us to conclude that no matter the database and descriptor used, there is a misleading improvement in the performance when using images of the same document both on training and testing sets.

Table 3: Identification rates (%) using LBP descriptor

| Database | # Writers | Without DF ($\sigma$) | With DF ($\sigma$) | DIF |
|---|---|---|---|---|
| $IAM_{v2}$ | 301 | 64.78 (±6.98) | 57.37 (±8.66) | 7.41 |
| BFL | 315 | 86.13 (±7.33) | 60.84 (±7.31) | 25.20 |
| CVL | 310 | 71.13 (±12.52) | 45.77 (±15.32) | 25.36 |

Table 4: Identification rates (%) using LPQ descriptor

| Database | # Writers | Without DF ($\sigma$) | With DF ($\sigma$) | DIF |
|---|---|---|---|---|
| $IAM_{v2}$ | 301 | 68.27 (±6.77) | 58.07 (±5.62) | 10.21 |
| BFL | 315 | 81.80 (±6.38) | 50.37 (±5.69) | 31.50 |
| CVL | 310 | 78.92 (±13.70) | 52.99 (±16.12) | 25.93 |

By analyzing the results presented in Tables 3 and 4, it is possible to conclude that there is a bias when data taken from different parts of the same document are used for training and test simultaneously. It can be easily observed in all the evaluated databases, and with all the features investigated as well. It was possible to note that the difference in performance can vary greatly among the different databases investigated, and also between the different features evaluated. Experiments on the IAM database using blocks taken from the original manuscript and the SURF descriptor achieved interesting results.

However, when texture blocks, and LBP and LPQ descriptors were used, the rates dropped. Apparently, the SURF descriptor is more robust on the IAM database than the other descriptors evaluated. Moreover, the SURF descriptor seems to be more suitable to the use of the DF protocol.

The best results on the BFL database were achieved using the LBP descriptor. On the CVL database, the three different descriptors provided results quite close to each other.

In order to contribute to a better understanding of the results described here, we will keep the same order in which the divisions of the dataset were organized and described in Section 3.2.

In the first case, the $IAM_{all}$ dataset was used, such a way that all the 657 writers were evaluated not using the DF protocol (i.e. "without DF"), since we have data taken from only a single manuscript from each writer in that database. In this case, what we really want to evaluate is the performance of a portion of the data that comprises all the writers, but all of them on a "without DF" mode. The identification rate obtained in this experiment was 70.12%.

On the second round of experiments, we check the writer identification performance on the IAM database using the DF protocol for every writer to which it is possible (i.e. authors with more than one manuscript sample). In this way, it was used two documents from the 301 writers who have more than one document, and one document from the other 356 writers, from whom there is only a single manuscript available. For the 301 writers we used blocks from the document 1 to test, and from the document 2 for training, or vice-versa. For the 356 writers, we used six blocks of the single manuscript available for training, and the tree remaining blocks for testing. It must not be forgotten that it is impossible to apply the DF protocol using all the 657 writers originally present in the IAM database. However, we performed this experiment to evaluate the performance using 657 writers from the IAM database, and taking into account the DF protocol as much as possible.

In this round we obtained an identification rate of 63.83%. The results obtained by now suggest that the use of DF tends to make the writer identification task harder then when it is not applied, since its usage has impacted the identification rate with a fall of 6.29%, what meets to our initial expectations.

Next, we describe results obtained using the $IAM_{v1}$ (i.e. 356 writers with a single manuscript). In this case, it is mandatory to use blocks taken from the same manuscript both on training and test sets. The experiment performed in this scenario, in which the DF protocol was never used, achieved a higher identification rate of 81.83%. Again, the obtained result confirms our preliminary expectations.

Until now, all the experiments have used single-sample writers (i.e. with the $IAM_{all}$ and $IAM_{v1}$ subsets), in the first and in the third round, only one document from all the writers was used. In the second round, we presented the results obtained using DF on part of the data. Now we describe the first experiment in which the DF protocol was used for every writer. In order to do that, we considered all the writers from whom we have two or more documents (i.e. we use the $IAM_{v2}$ subset). Since all writers in this experiment has at least two documents, we can evaluate two scenarios: i) with the DF protocol; and ii) without the DF protocol. The former scenario was addressed for obvious reasons, and the later was done aiming to compare the impacts of DF on exactly the same set of writers. The results are shown in the Table 5, and the difference in the identification rates whether or not using the DF protocol is noticeable, reaching 13.95%.

Table 5: Identification rates (%) using SURF descriptor

| Training | # Writers | Without DF ($\sigma$) | With DF ($\sigma$) | DIF |
|---|---|---|---|---|
| $IAM_{all}$ | 657 | 70.12 ($\pm$5.80) | - | - |
| $IAM_{v1}$ | 356 | 81.83 ($\pm$1.06) | - | - |
| $IAM_{v2}$ | 301 | 83.67 ($\pm$4.62) | 69.71 ($\pm$3.94) | 13.95 |

## 5  Critical review

In a first glance upon the problem addressed in this work, even those researchers used to the pattern recognition literature can think that this discussion is not applicable, since the use of different parts of the data taken from the same instance both in the training and testing sets may be a controversial practice, out of place in several other application domains. However, it is important to observe that this kind of fine tuning regarding the experimental protocol may vary widely from one application domain to another. As already mentioned here, Pampalk et al. [16] raised an intriguing question in 2005 about the results obtained until that moment in the music genre classification task. As a consequence, the authors introduced a new concept (i.e. artist filter) which have influenced most of the works in that field of research, having been considered as a restriction almost mandatory from that moment on.

In the case of the writer identification task, there are limits naturally imposed to the research community due to limitations of many databases typically used in this kind of work. This is why it is not particularly rare to find works concerning writer identification in which data taken from the same manuscript are used both on the training and test sets, what is inconceivable in other application domains.

There are some factors that can help to understand why the lack of the DF protocol can somehow introduce a bias to the results on writer identification. That is because of the same person has variations in its writing style in different occasions. In this regard, we can observe: the emotional state of the writer; the fluency of the pen/pencil used for writing; the lighting level around the writer; and the paper writability (smooth/rough), among others [12]. So, if we use fragments from the same manuscript on both training and testing sets, the identification rates tends to be biased.

That said, we should consider the feasibility of the use of the DF protocol when creating new manuscript databases, avoiding single-sample writers, since it could introduce some bias to the identification rates obtained using those data.

Finally, the last important matter to speculate here, regards to alternatives to minimize the impacts of single-sample writers on the development of experiments performed on databases already available in which it occurs. In this sense, we would conjecture that the use of data augmentation techniques could be tried aiming to mitigate the impact of single-sample writers. Otherwise, the researchers could consider to perform experiments applying the DF protocol on those parts of the data in which it is possible, like done here.

## 6 Concluding remarks

In this work we proposed a novel restriction to be imposed for the development of writer identification experiments, which we call "Document Filter" (DF). The DF protocol, enforce that all the data obtained from fragments of the same manuscript sample must be placed into one, and only one, set during the division to create training and test sets.

In this vein, we evaluate the impacts of the introduction of the DF protocol on three datasets widely used by the research community (i.e. IAM, BFL and CVL). Experiments showed that the identification rates tends to significantly decrease when the DF protocol is taken into account.

It is worth to mention that with this work we intend to show the effects of having single-sample writers when doing the writer identification task. We do not intend to discredit previous works or databases already described in the literature. In addition, we would suggest the enforcement of the DF protocol when creating/organizing databases in future, once in the most extreme cases of our experimentation the recognition rate obtained using the DF protocol dropped from 86.40% to 60.80% when it was used.

In future, we aim to investigate the impact of the availability of a restricted amount of text (i.e. only one paragraph per writer, only one line per writer, and only one word) on the performance of writer identification/verification systems.

## Acknowledgment

We thank the Brazilian research support agencies: Coordination for the Improvement of Higher Education Personnel (CAPES), and National Council for Scientific and Technological Development (CNPq) for their financial support.